\documentclass[lettersize,journal]{IEEEtran}

\usepackage{graphicx}
\usepackage{amsmath}
\usepackage{amssymb}

\usepackage[ruled,vlined]{algorithm2e}
\usepackage{hyperref}
\usepackage{float}
\usepackage{multirow}
\usepackage{booktabs,caption}
\usepackage[flushleft]{threeparttable}
\usepackage{subcaption}
\floatstyle{plaintop}
\restylefloat{table}

\usepackage{fancyhdr}
\usepackage{kantlipsum}
\usepackage{eso-pic}

\usepackage{color}
\usepackage{soul}

\begin{document}
\bstctlcite{IEEEexample:BSTcontrol}

\title{DeepIPCv3: Event-Aware Multi-Modal Sensor Fusion for\\Sudden Pedestrian Crossing Avoidance}

\author{Oskar Natan$^{1}$, Andi Dharmawan$^{1}$, Aufaclav Zatu Kusuma Frisky$^{1}$, Jazi Eko Istiyanto$^{1}$, and Jun Miura$^{2}$
	\thanks{Corresponding Author: Oskar Natan}
	\thanks{This research is funded by the Indonesian Endowment Fund for Education (LPDP) on behalf of the Indonesian Ministry of Higher Education, Science and Technology and managed under the EQUITY Program (Contract Number: 4301/B3/DT.03.08/2025 and 10107/UN1.P/Dit-Keu/HK.08.00/2025).}
	\thanks{$^{1}$Oskar Natan, Andi Dharmawan, Aufaclav Zatu Kusuma Frisky, and Jazi Eko Istiyanto are with the Department of Computer Science and Electronics, Universitas Gadjah Mada, Yogyakarta 55281, Indonesia. {\tt\small \{oskarnatan; andi\_dharmawan; aufaclav; jazi\}@ugm.ac.id}}%
	\thanks{$^{2}$Jun Miura is with the Department of Computer Science and Engineering, Toyohashi University of Technology, Toyohashi, Aichi 441-8580, Japan. {\tt\small jun.miura@tut.jp}}%
}

\markboth{THIS MANUSCRIPT HAS BEEN SUBMITTED TO IEEE FOR POSSIBLE PUBLICATION. COPYRIGHT MAY BE TRANSFERRED WITHOUT NOTICE.}%
{Natan \MakeLowercase{\textit{et al.}}: DeepIPCv3: Event-Aware Multi-Modal Sensor Fusion for Sudden Pedestrian Crossing Avoidance}


\maketitle

\begin{abstract}
	Current end-to-end autonomous driving systems predominantly rely on frame-based sensors, which suffer from inherent perception latency and motion blur during highly dynamic encounters, specifically sudden pedestrian crossings. To address this critical safety vulnerability, we propose DeepIPCv3, a novel multi-modal autonomous navigation framework that synergizes the dense 3D spatial geometry of LiDAR point clouds with the microsecond-level asynchronous event streams of a Dynamic Vision Sensor (DVS). We introduce a Transformer-inspired cross-modal attention mechanism to dynamically correlate these distinct modalities, allowing the network to instantaneously prioritize high-speed dynamic updates without sacrificing structural scene awareness. The fused latent representations are then mapped to safe local waypoints and executable control commands via a hybrid policy network that blends heuristic trajectory tracking with direct neural predictions. Due to the severe physical risks associated with live testing of these sudden-crossing scenarios, the framework is rigorously evaluated offline using a custom multi-modal dataset collected across both well-illuminated noon and challenging evening conditions. Extensive comparative and ablation studies demonstrate that DeepIPCv3 achieves state-of-the-art predictive performance. By effectively eliminating exposure failures and motion blur, the proposed LiDAR and DVS fusion yields the lowest trajectory and control command errors, enabling highly reactive, mathematically bounded evasive maneuvers regardless of ambient illumination. To support future research, we will release the codes to our GitHub repo at \href{https://github.com/oskarnatan/DeepIPCv3}{https://github.com/oskarnatan/DeepIPCv3}.
\end{abstract}

\begin{IEEEkeywords}
	Autonomous Vehicles, Visual Perception, Sensor Fusion, Multi-Modal Learning, Reactive Navigation.
\end{IEEEkeywords}

\section{Introduction}\label{sec:intro}
\IEEEPARstart{E}{nd}-to-end autonomous driving has emerged as a compelling paradigm for outdoor point-to-point navigation, offering a streamlined alternative to traditional, heavily modularized pipelines \cite{e2etvt}\cite{legodrive}. By directly mapping raw multi-modal sensor inputs to navigational waypoints and control actions, these architectures reduce compounding errors and simplify system deployment \cite{e2evtc}\cite{oskar_tiv}. Recent advancements leveraging high-fidelity sensors, such as RGB-D cameras and 3D LiDAR, have significantly improved the robustness of environmental perception, semantic scene understanding, and trajectory prediction under standard operating conditions \cite{multimodalav}\cite{lidarcampercep}. However, ensuring consistent and safe navigation in unstructured, dynamic outdoor environments remains a formidable challenge, particularly when the vehicle must balance goal-directed routing with instantaneous obstacle avoidance \cite{safedriving}\cite{safedriving2}.

\begin{figure}
	\begin{center}
		\includegraphics[width=\linewidth]{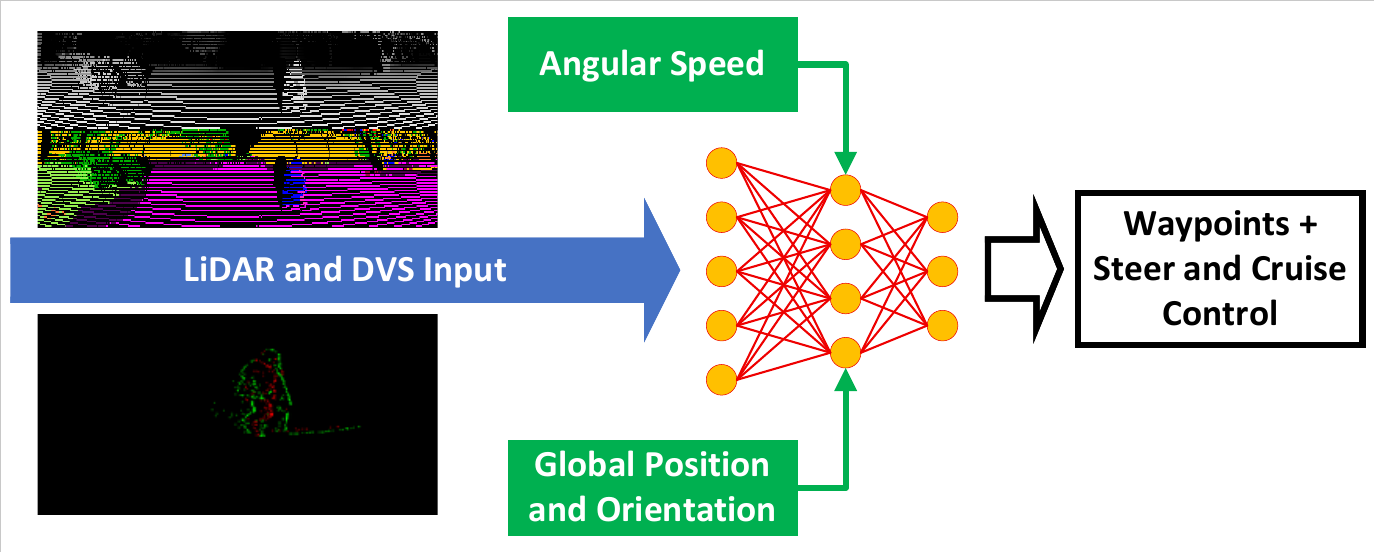}
	\end{center}
	\vspace{-7pt}
	\caption{DeepIPCv3 observes the environment with LiDAR and DVS. Then, predict a set of navigational outputs.}
	\label{fig:graphabs}
	\vspace{-10pt}
\end{figure}

While current approaches successfully fuse spatial geometry with temporal modeling to navigate static and predictable environments, they frequently struggle under highly dynamic conditions \cite{interfuser}\cite{transfuser}. A critical safety scenario that remains inadequately addressed is the sudden, unpredictable appearance of dynamic obstacles \cite{penaltyav}\cite{dynamicobs}. In this study, we specifically focus on sudden dynamic obstacle encounters, specifically a pedestrian rapidly crossing the vehicle's path. Conventional frame-based sensors suffer from inherent latency and motion blur during fast movements, which bottlenecks the system's reaction time \cite{lmdrive}\cite{aim_mt}\cite{huang_model}. Although event-based vision like the Dynamic Vision Sensor (DVS) offers microsecond-level temporal resolution by asynchronously capturing localized pixel-intensity changes \cite{dvssens}\cite{dvssens2}, effectively integrating these sparse, asynchronous event streams with dense 3D point clouds remains an open research problem.

To bridge this gap, we propose DeepIPCv3 as shown in Fig. \ref{fig:graphabs}, an upgrade of our prior works \cite{deepipc}\cite{deepipcv2}, a novel end-to-end autonomous driving framework that integrates LiDAR-based spatial perception with DVS event streams to achieve robust point-to-point navigation and ultra-low-latency responses to execute safe avoidance maneuvers. Recognizing the distinct data modalities of 3D point clouds and event-based intensity changes, we introduce a sophisticated data fusion mechanism driven by Transformer-inspired attention blocks to dynamically learn cross-modal relationships. Rather than deploying the system in a high-risk online environment for these sudden-crossing scenarios, the framework is rigorously evaluated offline. DeepIPCv3 focuses strictly on learning the accurate policy mapping from the raw, fused sensor data to precise control commands. The main contributions of this work are summarized as follows:
\begin{itemize}
	\item We propose a multi-modal architecture that synergizes LiDAR and DVS inputs, enabling the system to maintain stable point-to-point navigation while improving responsiveness to sudden dynamic obstacles.
	\item We introduce a Transformer-inspired cross-modal fusion module, which can extracts and correlates features between dense 3D spatial geometry and asynchronous event streams to generate optimal driving policies.
	\item We conduct extensive offline evaluations using a custom multi-modal dataset featuring sudden pedestrian crossings under both well-illuminated and illumination-degraded conditions. The results demonstrate that DeepIPCv3 achieves state-of-the-art predictive accuracy, effectively decoupling dynamic responsiveness from ambient lighting to execute contextually safe avoidance behaviors, such as evasive steering or full-stop braking.
\end{itemize}

\section{Related Works}

In this section, we do literature review on some related works that focus on end-to-end autonomous driving in sudden dynamic encounters and event-based autonomous navigation. Then, we point out some remain issues and also highlight key ideas for comparative study.

\subsection{Autonomous Driving in Sudden Dynamic Encounters}
The shift toward end-to-end autonomous driving has streamlined the navigation pipeline, yet maintaining robustness under highly dynamic conditions remains a challenge \cite{dynobs}\cite{obs2}. For instance, research into dynamic deception demonstrates that end-to-end models heavily reliant on frame-based perception can be easily compromised by pedestrians suddenly crossing the vehicle's path. Recent studies leveraging high-fidelity simulations, such as CARLA \cite{carla}, have increasingly focused on the vulnerability of these systems to sudden dynamic obstacles \cite{leapvad}\cite{carlabrin}\cite{neat}. However, standard vision sensors operate at fixed frame rates suffer from inherent latency and motion blur during rapid, unpredicted environmental changes \cite{illum}. To mitigate this, several recent frameworks have attempted to append physics-informed safety controllers \cite{phyinform} or model predictive control (MPC) modules \cite{mpcav} to the end-to-end pipeline. While these hybrid approaches can enforce hard safety constraints to prevent collisions, they often do so at the cost of the system's fluidity, reacting aggressively rather than proactively due to the delayed perception of the fast-moving obstacle. 

Despite these algorithmic advancements, the fundamental bottleneck remains the sensory latency during rapid dynamic events, which algorithmic safety patches cannot entirely eliminate. Rather than relying solely on post-perception control overrides, DeepIPCv3 addresses this vulnerability at the sensory root by integrating an asynchronous event camera, so that it can fundamentally reduces perception latency, natively capturing the high-speed motion of sudden pedestrian crossings before frame-based sensors can register the blur.

\subsection{Event-Based Autonomous Navigation}
To overcome the limitations of traditional frame-based cameras, event-based vision has gained significant traction in the autonomous driving domain \cite{eddd}\cite{mimooskar}. Dynamic Vision Sensors (DVS) operate differently from standard cameras by asynchronously recording only localized changes in pixel intensity, thereby providing microsecond-level temporal resolution and an exceptionally high dynamic range \cite{dvsreview}\cite{speeddvs}. Pioneering datasets such as DDD20 \cite{ddd20} and EvTTC \cite{evttc} have catalyzed the development of deep learning models, including brain-inspired liquid neural networks and specialized convolutional recurrent networks, that directly map event streams to various driving tasks. Recent architectures have also demonstrated the efficacy of event cameras in overcoming motion blur and varying illumination during continuous driving tasks \cite{eve1}\cite{eve2}. These models show that event streams can drastically improve reaction times and maintain consistent performance across massive domain shifts, which frequently confounds standard RGB cameras. 

However, while existing event-based models excel at reactive steering, they often lack the dense 3D spatial awareness. We overcomes this limitation by synergizing the DVS event streams with the precise 3D geometric mapping of LiDAR point clouds. DeepIPCv3 leverages a Transformer-inspired cross-modal attention mechanism to dynamically weigh these inputs, extracting the spatial context from the LiDAR while maintaining the responsiveness of the DVS.

\section{Proposed Methods}

In this section, we first describe the problem formulation of this research. Then, we explain the model in detail, especially on how the perception and controller modules work. We also describe the dataset used to train, validate, and test the model. Finally, we explain the loss function formulation and the training configuration.

\begin{figure*}
	\begin{center}
		\includegraphics[width=\linewidth]{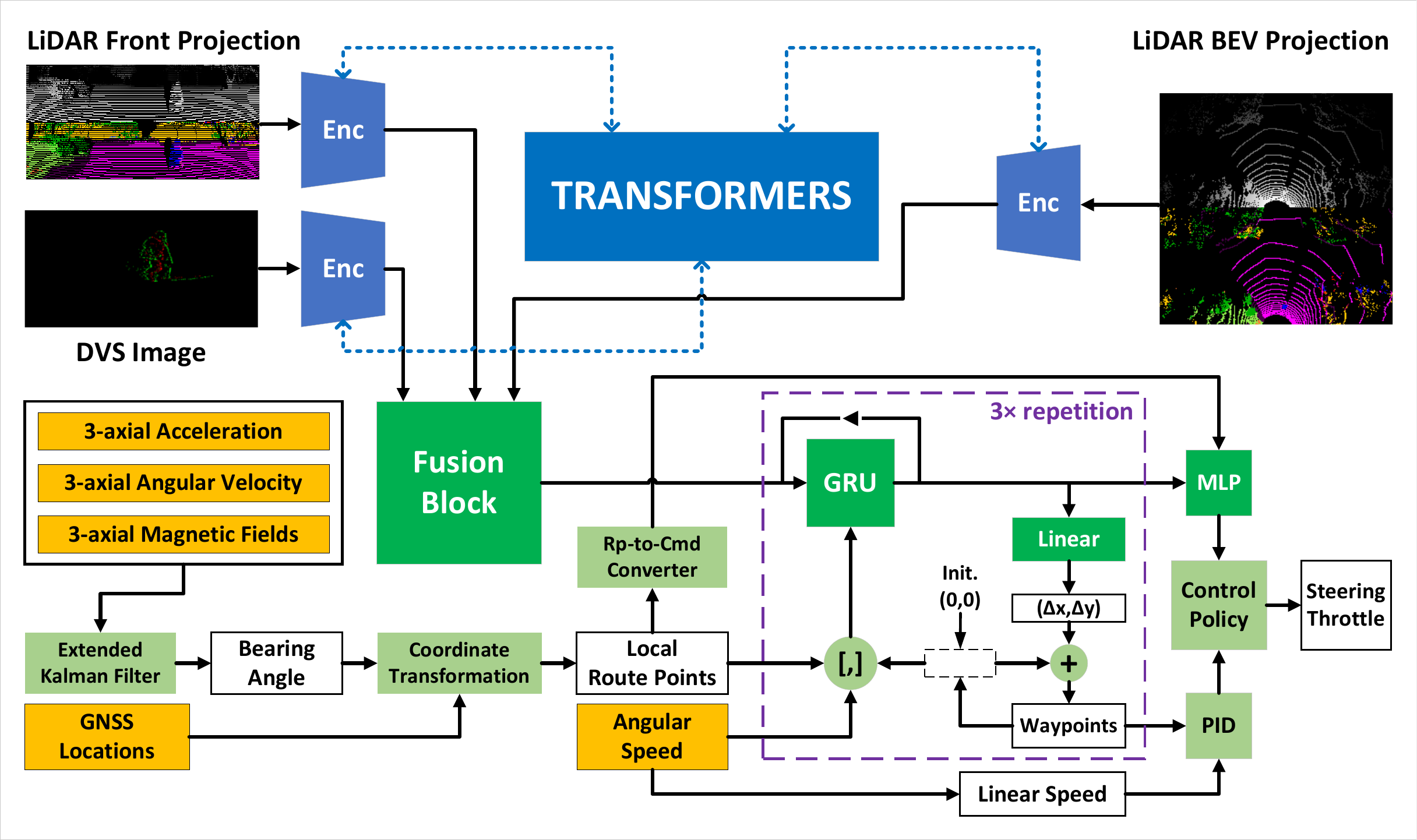}
	\end{center}
	\vspace{-7pt}
	\caption{The architecture of DeepIPCv3. Blue blocks are considered as part of the perception module, while green blocks are considered as part of the planning-control module. Light-colored blocks are not trainable, they are fixed functions.}
	\label{fig:model}
	\vspace{-5pt}
\end{figure*}

\subsection{Problem Formulation}
The autonomous navigation task is formulated as a multimodal imitation learning problem within a dynamic observable environment. At any given timestamp $t$, the ego-vehicle receives a comprehensive observation tuple $\mathcal{O}_t = \{ L_t, R_t, E_{t-\Delta t}^{t}, N_t \}$. This tuple represents the dense 3D LiDAR point cloud ($L_t$), the RGB-D spatial frame ($R_t$), the continuous asynchronous DVS event stream ($E_{t-\Delta t}^{t}$) accumulated over a micro-temporal window $\Delta t$, and the vehicle's proprioceptive navigational state ($N_t$). $N_t = \{ c_t, \omega_t, \psi_t \}$ comprises the latitude-longitude coordinates $c_t$ obtained from the GNSS receiver, the angular speed $\omega_t$ from the rotary encoders, and the global orientation $\psi_t$ from the 9-axis IMU. This data is strictly required to generate and transform sparse global route points into ego-centric local route points to achieve goal-directed point-to-point navigation.

The fundamental objective is to learn an optimal end-to-end driving policy, parameterized by a deep neural network $\pi_\theta$, that directly maps the high-dimensional observation space to a safe, executable action space. It is defined as $\mathcal{A}_t = \{ \mathcal{W}_t, U_t \}$, encompassing a sequence of $M$ future local waypoints $\mathcal{W}_t = \{(x_i, y_i)\}_{i=1}^{M}$ in the ego-coordinate frame, alongside the explicit control commands $U_t \in \{\text{steer, cruise}\}$. Formally, let the expert demonstration dataset be defined as $\mathcal{D} = \{ (\mathcal{O}_t, \mathcal{Y}^*_t) \}_{t=1}^{T}$, where $\mathcal{Y}^*_t = \{ \mathcal{W}^*_t, U^*_t \}$ represents the ground-truth waypoints and control actions executed by a human expert. The overarching goal is to determine the optimal network parameters $\theta^*$ that minimize the expected imitation loss $\mathcal{L}$ over the empirical data distribution $p_{data}$:

\begin{equation}
	\theta^* = \arg\min_{\theta} \mathbb{E}_{(\mathcal{O}, \mathcal{Y}^*) \sim p_{data}} \left[ \mathcal{L} \left( \pi_\theta(\mathcal{O}), \mathcal{Y}^* \right) \right].
\end{equation}

Keep in mind that the dataset contains observation-action pairs representing sudden dynamic encounters. Specifically, $\mathcal{O}$ captures sudden pedestrian crossings, and $\mathcal{Y}^*$ contains the corresponding safe avoidance actions executed by the expert driver. Thus, the proposed policy $\pi_\theta$ must not only minimize the static trajectory tracking error over the entire route but also satisfy a strict instantaneous reactivity constraint during these dynamic encounters. Formally, if an expert initiates an avoidance maneuver at timestamp $t_{expert}$, the learned policy must execute the corresponding command at $t_{pred}$ such that the temporal lag, defined as $t_{pred} - t_{expert}$, approaches zero. It is crucial to note that this temporal lag is not explicitly penalized within the training loss formulation, as providing direct supervisory cues regarding the exact timing of sudden events would compromise the model's generalized reactivity. Instead, minimizing this lag serves as the fundamental architectural objective, achieved implicitly by prioritizing the low-latency event streams to replicate the expert's rapid evasive maneuvers prior to the onset of frame-based motion blur. We would like the near-zero lag to be an emergent property of the sensor fusion rather than a forced optimization target.


\begin{figure*}
	\begin{center}
		\includegraphics[width=\linewidth]{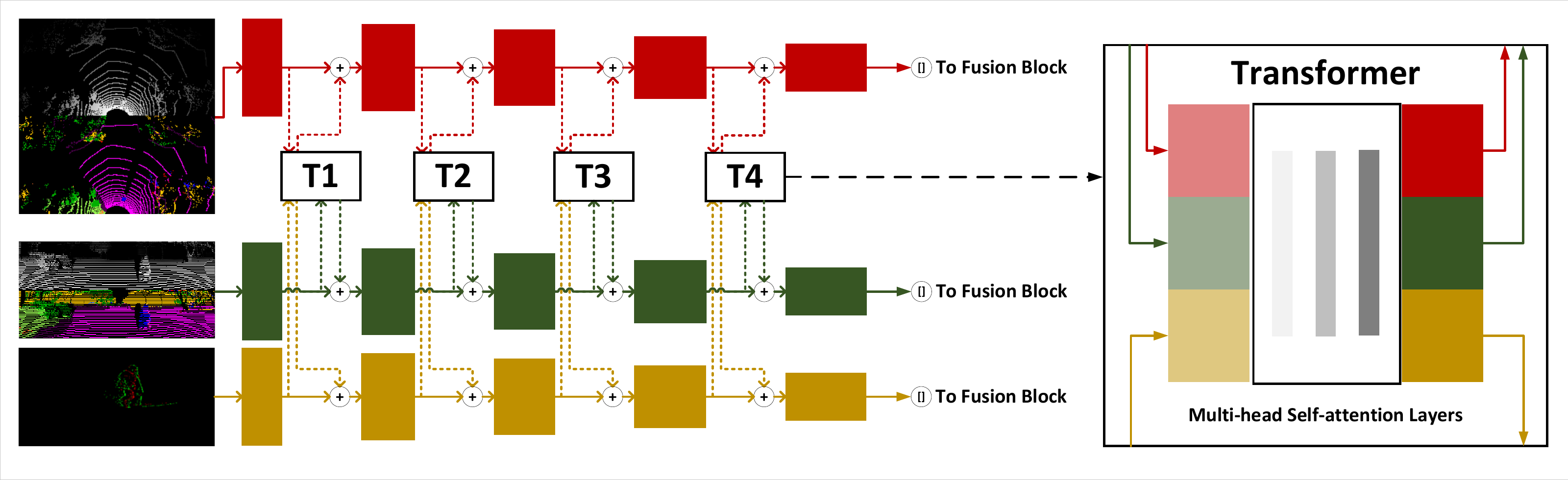}
	\end{center}
	\vspace{-7pt}
	\caption{The architecture of the transformer attention blocks.}
	\label{fig:transformer_module}
	\vspace{-5pt}
\end{figure*}

\subsection{DeepIPCv3 Architecture} \label{subs:model}
To achieve robust autonomous navigation under both static and highly dynamic conditions, we propose DeepIPCv3. As illustrated in Fig. \ref{fig:model}, DeepIPCv3 architecture is divided into two primary pipelines: the multi-modal perception module and the planning-control module. To effectively synthesize the disparate feature maps, we introduce a dedicated data fusion module based on Transformer-inspired attention blocks \cite{interfuser}\cite{attn}, detailed in Fig. \ref{fig:transformer_module}. Unlike simple concatenation or temporal recurrent fusion, the Transformer module  employs self-attention and cross-attention mechanisms to learn the complex relationships between the input modalities. This allows the network to dynamically shift its "attention" depending on the environmental context. Then, the unified latent representation is passed to the planning-control module. Rather than relying on rigid, post-processing heuristic controllers, this module acts as a sophisticated policy network. It learns the direct policy mapping from the raw, fused sensor data to precise control commands. Specifically, the network predicts safe local waypoints and translates them into explicit high-level commands. By isolating the predictive intelligence to output these exact positional and orientational targets, the lower-level actuation can be efficiently handled by the vehicle's separate kinematics system.

\subsubsection{Multi-Modal Perception and Transformer-Based Fusion}

To achieve robust autonomous navigation, DeepIPCv3 ingests two distinct sensory modalities: 3D LiDAR point clouds and asynchronous event streams from the Dynamic Vision Sensor (DVS). The raw asynchronous DVS events stream within $\Delta t$ are aggregated into a 3D spatio-temporal voxel grid to form a dense tensor. We also consider RGB-D images as the input for the ablation study. Similar to DeepIPCv2 \cite{deepipcv2}, the proposed model leverages PolarNet \cite{polarnet} pretrained on KITTI dataset \cite{kitti} for efficient point cloud segmentation to construct a robust spatial prior. Then, we concatenate both segmentation and depth maps projected from the point clouds. To extract high-dimensional spatial and temporal features without introducing excessive computational overhead, we employ EfficientNet-B0 \cite{efficientnet} as the backbone architecture for our parallel encoders. Let the extracted flattened feature sequences be denoted as $X_L \in \mathbb{R}^{N \times D}$ for LiDAR, $X_D \in \mathbb{R}^{N \times D}$ for the DVS event streams, and $X_R \in \mathbb{R}^{N \times D}$ for RGB-D (ablation study), where $N$ is the sequence length and $D$ is the embedding dimension. Standard concatenation of these feature maps often fails to capture the dynamic interplay between modalities, especially when a sudden pedestrian crossing renders the frame-based $X_R$ obsolete due to motion blur. Therefore, we introduce a cross-modal Transformer fusion module to dynamically weigh these inputs, allowing the network to focus computational capacity on dynamic obstacle avoidance and policy mapping. Specifically, the dense spatial geometry from the LiDAR ($X_L$) acts as the spatial anchor (Query), while the DVS event features ($X_D$) provide the instantaneous dynamic updates (Key and Value). The cross-attention mechanism is mathematically formulated as:

\begin{equation}
	Q = X_L W_Q, \quad K = X_D W_K, \quad V = X_D W_V,
\end{equation}
where $W_Q, W_K, W_V \in \mathbb{R}^{D \times d_k}$ are the learnable linear projection matrices. The cross-attended latent representation $Z_{LD}$ is then computed as:

\begin{equation}
	Z_{LD} = \text{Softmax}\left(\frac{Q K^T}{\sqrt{d_k}}\right)V,
\end{equation}
where $\sqrt{d_k}$ acts as a scaling factor to prevent vanishing gradients. This process is repeated across multi-head attention blocks, yielding a comprehensively fused and context-aware latent vector, $Z_{fused}$, which serves as the foundation for the subsequent control policy. %

\subsubsection{Waypoint Prediction and Hybrid Control Formulation}

The planning-control module translates the context-aware latent representations from the perception module into safe navigational waypoints and executable vehicular commands. As illustrated by the "Fusion Block" in Fig. \ref{fig:model}, the multi-modal features extracted by the encoders and the Transformer are flattened and aggregated into a single visual latent vector, $Z_{fused}$. It should be noted that $Z_{fused}$ encapsulates only the exteroceptive environmental features. To achieve goal-directed navigation, this vector is subsequently passed to a Gated Recurrent Unit (GRU) \cite{gru} loop, where it is concatenated with the vehicle's proprioceptive navigational state vector $N_t$ (which includes local route points and angular speed). Then, a dedicated multi-layer perceptron (MLP) branch regresses a series of local target waypoints $\mathcal{W} = \{(x_i, y_i)\}_{i=1}^{M}$ in the vehicle's ego-coordinate system. To govern the vehicle's physical movement, we adopt a hybrid control strategy that combines traditional proportional-integral-derivative (PID) tracking with direct neural predictions. This planning-control component is kept identical to the module utilized in our previous architectures, DeepIPC, \cite{deepipc} DeepIPCv2, and \cite{deepipcv2} Seq-DeepIPC \cite{seqdeepipc}, as it has proven highly stable. Thus, the architectural upgrades in DeepIPCv3 focus exclusively on the perception and multi-modal fusion stages. The PID controller translates the immediate predicted waypoint $(x_1, y_1)$ into high-level heuristic control targets. For instance, the orientation error $e_\theta(t)$ is calculated based on the target heading $\theta_{target} = \arctan(y_1 / x_1)$ relative to the vehicle's current heading. The heuristic steer command $u_{PID}^{steer}$ is given by:

\vspace{5pt}
\begin{equation}
	u_{PID}^{steer}(t) = K_p e_\theta(t) + K_i \int_{0}^{t} e_\theta(\tau) d\tau + K_d \frac{de_\theta(t)}{dt}.
\end{equation}

A similar PID formulation computes the heuristic cruise control $u_{PID}^{cruise}$ based on the longitudinal distance to the waypoints. Simultaneously, a separate command-specific MLP decodes $Z_{fused}$ to directly predict the required control adjustments $u_{MLP}$, capturing complex, non-linear driving behaviors (such as hard braking for a sudden pedestrian) that standard PID logic might smooth over. The final executed control commands, $U_{final} \in \{\text{steer, cruise}\}$, are formulated as a weighted combination of both the heuristic PID outputs and the neural predictions:

\begin{equation}
	U_{final} = \alpha \cdot u_{PID} + (1 - \alpha) \cdot u_{MLP},
\end{equation}
where $\alpha \in [0, 1]$ is a learnable gating parameter tuned by the Modified Gradient Normalization (MGN) algorithm \cite{mgn} as in our prior works \cite{deepipc}\cite{deepipcv2}. This hybrid architecture ensures that the system maintains smooth, mathematically bounded trajectory tracking via the PID, while the MLP enables instantaneous maneuvers in sudden dynamic encounters.

\subsection{Dataset Collection}
Building upon the formalized problem statement, the empirical dataset $\mathcal{D}$ was collected in a structured outdoor environment at Toyohashi
University of Technology, Japan, featuring a mix of straight navigational corridors, intersections, and dynamic obstacles similar to our prior works \cite{deepipc}\cite{deepipcv2}. To ensure the policy network $\pi_\theta$ learns to handle the critical temporal constraints associated with sudden pedestrian crossings, the data collection protocol deliberately included a high frequency of challenging scenarios where the pedestrian's trajectory intersects the vehicle's path. This data was captured under two distinct environmental conditions: well-illuminated noon and illumination-degraded evening settings.

In total, the dataset consists of 18 distinct route trajectories recorded at a sampling rate of 4 Hz, yielding a total of 37,855 synchronized multi-modal frames. The unique routes are partitioned into standard training, validation, and testing splits with a ratio of 6:6:6. Quantitatively, the training set contains 9,033 frames (4,767 noon; 4,266 evening), and the validation set contains 8,975 frames (4,200 noon; 4,775 evening). To vary the dynamic scenarios and environmental conditions for a better evaluation, data for the strictly unseen testing subset, $\mathcal{D}_{test}$, was gathered by traversing its 6 designated routes three separate times on different days. Consequently, the testing subset is larger than the training and validation sets, comprising 19,847 frames (9,972 noon; 9,875 evening). The training and validation subsets are utilized to optimize the network weights, while the testing subset is reserved for offline evaluations. This split ensures that the model's reported performance reflects its true capability to generalize across both steady-state navigation and reactive evasive maneuvers. To visually demonstrate the difficulty of these sudden pedestrian-crossing scenarios, supplementary video materials have been made available on our GitHub repository.

\subsection{Loss Functions and Training Configuration} \label{subs:train}

DeepIPCv3 is implemented with PyTorch Framework \cite{torch} trained in an end-to-end manner using a multi-task learning paradigm that jointly optimize both planning and control objectives. The overarching function integrates specific loss components corresponding to the navigational outputs. Specifically, the local waypoint prediction loss $\mathcal{L}_{wp}$ and the control command loss $\mathcal{L}_{cmd}$ (representing the $x$-$y$ maneuverability or steer-cruise regressions), are formulated using the combination of $L_1$ loss and $L_2$ loss to penalize continuous regression deviations. The total multi-task loss $\mathcal{L}_{total}$ is formulated as the weighted sum of these individual task losses:

\begin{equation}
	\mathcal{L}_{total} = w_{wp}\mathcal{L}_{wp} + w_{cmd}\mathcal{L}_{cmd}
\end{equation}
where $w_{wp}$ and $w_{cmd}$ denote the respective task weights. Because static weight assignment in multi-task networks frequently leads to task dominance and sub-optimal convergence \cite{mtlissue}, we implement the Modified Gradient Normalization (MGN) algorithm \cite{mgn} to balance the loss weights adaptively. During backpropagation, the MGN algorithm dynamically scales the task-specific gradients, ensuring a balanced optimization landscape and allowing the shared EfficientNet backbones and Transformer fusion modules to generalize equally well across all objectives. To be noted, for the comparative study models that explicitly predict intermediate vision tasks (e.g., semantic segmentation or depth estimation), we also define a perception loss $\mathcal{L}_{perc}$ as in their papers.

The network parameters are optimized using Adam with decoupled weight decay, which provides superior regularization and helps prevent overfitting on the highly dynamic datasets \cite{adamw}. To ensure stable convergence, the training process is governed by a dynamic learning rate, which is reduced by halved if there is no improvement in the validation loss for 5 consecutive epochs. Similarly, an early stopping mechanism is integrated to terminate the training process if there is no improvement for 20 consecutive epochs.

\begin{table*}
	\vspace{3pt}
	\caption{Model Specification}
	\begin{center}
		\resizebox{\textwidth}{!}{%
			\begin{tabular}{ccccc}
				\toprule
				Model & Total Parameters$\downarrow$ & Model Size $\downarrow$ & Input/Sensor & Output\\
				\toprule
				Huang \textit{et al.} \cite{huang_model} & 74.95 M & 300.20 MB & RGBD, High-level commands & Segmentation, Steer, Cruise\\
				AIM-MT \cite{aim_mt} & 27.97 M & 112.08 MB & RGB, GNSS, 9-axis IMU, Rotary encoder & Segmentation, Depth, Waypoints, Steer, Cruise\\
				LMDrive \cite{lmdrive} & 31.32 M & 122.45 MB & RGB, High-level commands & Waypoints, Steer, Cruise\\
				TransFuser \cite{transfuser} & 66.23 M & 265.38 MB & RGB, LiDAR, GNSS, 9-axis IMU, Rotary encoder & Waypoints, Steer, Cruise\\
				
				DeepIPC \cite{deepipc} & 20.98 M & 84.97 MB & RGBD, GNSS, 9-axis IMU, Rotary encoder & Segmentation, Waypoints, Steer, Cruise\\
				DeepIPCv2 \cite{deepipcv2} & 19.95 M & 78.54 MB & LiDAR, GNSS, 9-axis IMU, Rotary encoder & Waypoints, Steer, Cruise\\
				DeepIPCv3 & 22.43 M & 86.60 MB & LiDAR, DVS, GNSS, 9-axis IMU, Rotary encoder & Waypoints, Steer, Cruise\\
				\bottomrule                             
			\end{tabular}
		}
	\end{center}
	\label{tab:model_compare}
	\vspace{-7pt}
\end{table*}

\section{Experimental Setup}
To validate the proposed DeepIPCv3 architecture, particularly its responsiveness in sudden dynamic scenarios, we conducted extensive offline evaluations. This section details the quantitative metrics used to assess the network's predictive performance, alongside the comparative baselines and the specific ablation studies designed to isolate the contributions of the proposed multi-modal fusion.

\subsection{Testing and Evaluation Metrics}
Because evaluating the sudden pedestrian-crossing scenarios in a live, online environment presents severe safety risks, DeepIPCv3 is strictly evaluated offline using a dedicated test dataset, $\mathcal{D}_{test}$, which also contains unseen standard and sudden pedestrian crossing sequences. The evaluation focuses on quantifying the accuracy of the learned policy mapping from raw sensor inputs to both local waypoints and high-level control commands.

To evaluate the spatial accuracy of the predicted local waypoints $\mathcal{W} = \{(x_i, y_i)\}_{i=1}^{M}$ against the expert ground truth $\mathcal{W}^* = \{(x^*_i, y^*_i)\}_{i=1}^{M}$, we employ the Average Displacement Error (ADE) that measures the mean Euclidean distance across all predicted waypoints in the sequence, which is highly indicative of the model's predictive stability:

\begin{figure}[t]
	\begin{center}
		\includegraphics[width=\linewidth]{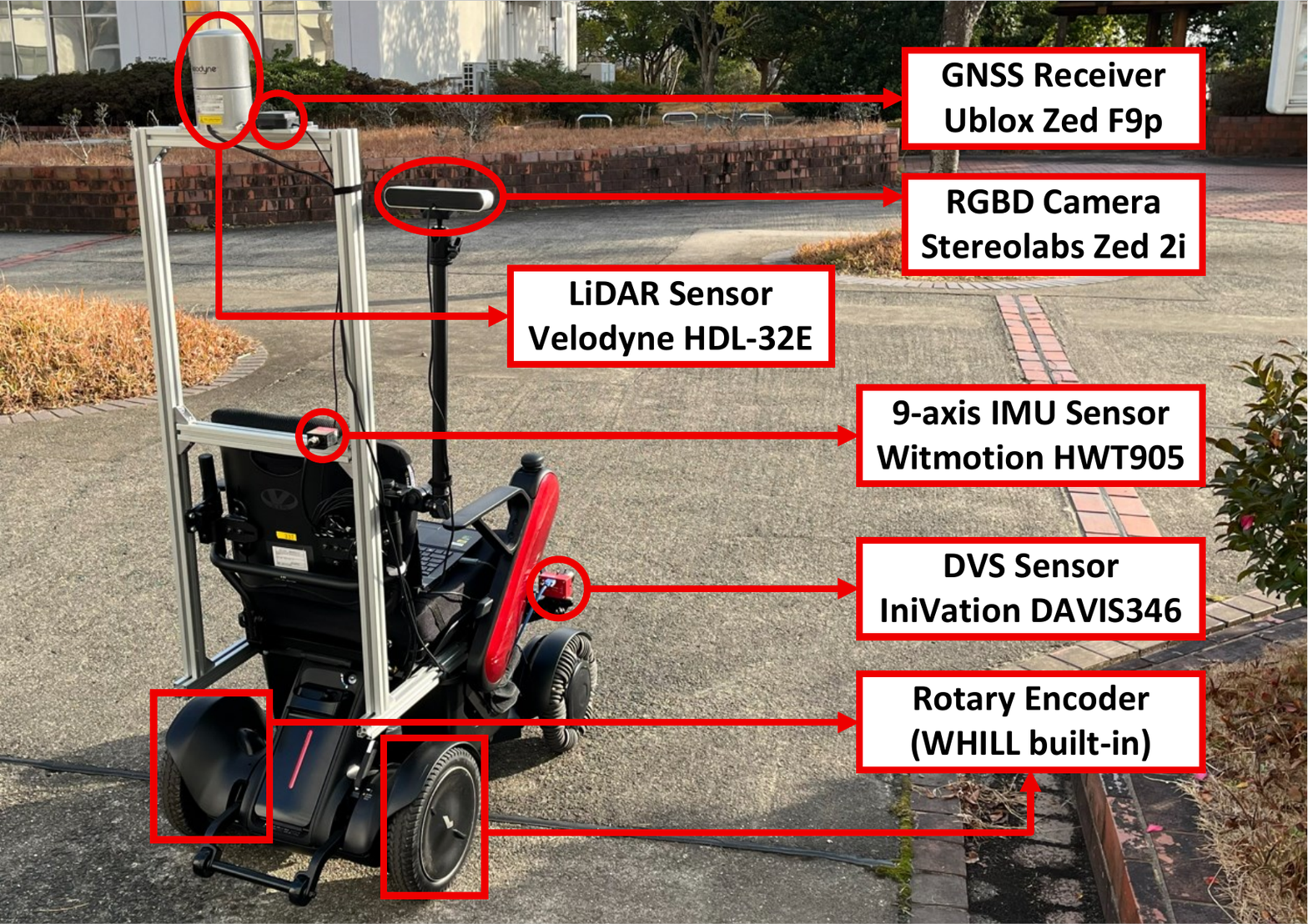}
		
	\end{center}
	\caption{Sensors and robot setup.} 
	\label{fig:sensor_setup}
	\vspace{-10pt}
\end{figure}

\begin{equation}
	\text{ADE} = \frac{1}{M} \sum_{i=1}^{M} \sqrt{(x_i - x^*_i)^2 + (y_i - y^*_i)^2}.
\end{equation}

For the control policy evaluation, we quantify the deviation of the final executed commands, $U_{final} \in \{\text{steer, cruise}\}$, against the expert control commands $U^*$. This is measured using the Mean Absolute Error (MAE):

\begin{equation}
	\text{MAE}_{cmd} = \frac{1}{N_{test}} \sum_{j=1}^{N_{test}} \left| U_{final}^{(j)} - U^{*(j)} \right|,
\end{equation}
where $N_{test}$ represents the total number of evaluation frames in $\mathcal{D}_{test}$. Furthermore, to explicitly quantify the model's responsiveness during sudden dynamic encounters, we introduce the Response Time Delay (RTD) metric. Since the dataset is recorded at 4 Hz, each frame sequence advances in 250 ms increments. RTD measures the temporal lag (in seconds) between the expert driver's initiation of the avoidance maneuver (e.g., the exact timestamp where hard braking or sharp steering commences) and the model's corresponding predicted maneuver:

\begin{equation}
	\text{RTD} = \frac{1}{K} \sum_{k=1}^{K} \left( t_{pred}^{(k)} - t_{expert}^{(k)} \right)
\end{equation}
where $K$ is the total number of sudden pedestrian encounters in $\mathcal{D}_{test}$, $t_{expert}$ is the timestamp of the expert's initial evasive reaction, and $t_{pred}$ is the timestamp when the model's predicted control command surpasses the same reaction threshold. In this study, the reaction threshold is quantitatively defined as the moment the absolute frame-to-frame change in the control signal ($|U_{t} - U_{t-1}|$) exceeds a predefined baseline margin $\epsilon$. This signifies a deliberate, sharp departure from steady-state navigation, such as an aggressive drop in the cruise command (hard braking) or a sudden spike in the steering value (evasive swerve). A lower RTD indicates faster responsiveness, with 0.0 seconds representing perfect synchronization with the expert.

\subsection{Robot and Sensor Configuration}
To evaluate the proposed framework in real-world environments and collect the offline dataset $\mathcal{D}$, we instrumented a WHILL Model C2 mobile robot. The robot serves as an agile and highly maneuverable platform suitable for navigating pedestrian-heavy outdoor environments. For proprioceptive state estimation and localization, the vehicle is equipped with a U-blox ZED-F9P GNSS receiver to capture global positioning, a Witmotion HWT905 9-axis IMU sensor to measure high-frequency orientation, and the WHILL's internal rotary encoders to record the vehicle's angular speed.

The exteroceptive perception payload is strategically mounted to provide overlapping, multi-modal fields of view, as illustrated in Fig. \ref{fig:sensor_setup}. To capture the dense 3D spatial geometry required for structural scene understanding, a Velodyne HDL-32E LiDAR is mounted centrally to provide a 360-degree point cloud, $L_t$. To capture the high-speed dynamic updates necessary for the sudden pedestrian-crossing scenarios, an Inivation DAVIS 346 Dynamic Vision Sensor (DVS) is mounted forward-facing, capturing asynchronous event streams $E_t$ with microsecond temporal resolution. Furthermore, a Stereolabs ZED 2 RGB-D camera is positioned alongside the DVS. While the DeepIPCv3 core architecture primarily relies on the LiDAR and DVS fusion, the RGB-D camera actively records synchronized spatial frames, $R_t$. Retaining the ZED 2 in the sensor suite ensures a rigorously matched multi-modal dataset, enabling fair, direct comparative baselines and comprehensive ablation studies against earlier architectures.

\section{Results and Discussions}

Due to the severe safety risks associated with testing these sudden-crossing scenarios in a live environment where sudden pedestrian crossings could result in damage to both the pedestrians and the physical vehicle, all evaluations were strictly conducted offline. All models mentioned in Table \ref{tab:model_compare} were tasked with predicting navigational outputs across a dedicated test dataset ($\mathcal{D}_{test}$), which includes highly dynamic sequences recorded during both noon and evening conditions. To ensure statistical reliability, every model variant and baseline was trained and evaluated three independent times, with the results reported as the average and deviation across these runs.

\begin{table*}[htbp]
	\vspace{3pt}
	\centering
	\caption{Ablation Results}
	\label{tab:abl_results}
	\resizebox{\textwidth}{!}{%
		\begin{tabular}{lll cccc}
			\toprule
			\multirow{2}{*}{\textbf{Condition}} &
			\multirow{2}{*}{\textbf{Variant}} & \multirow{2}{*}{\textbf{Perception Input}} &  \textbf{Planning Task} & \multicolumn{3}{c}{\textbf{Control Task}} \\
			\cmidrule(lr){4-4} \cmidrule(lr){5-7}
			& & & \textbf{Waypoints ADE $\downarrow$} & \textbf{Steer MAE $\downarrow$} & \textbf{Cruise MAE $\downarrow$} & \textbf{RTD (s) $\downarrow$} \\
			\midrule
			\multirow{6}{*}{Noon}
			& & LiDAR+RGB+DVS & 0.095 $\pm$0.004 & 0.119 $\pm$0.005 & 0.058 $\pm$0.003 & 0.231 $\pm$0.067  \\
			& w/o Transformer & LiDAR+RGB & 0.103 $\pm$0.004 & 0.123 $\pm$0.004 & 0.055 $\pm$0.003 & 0.238 $\pm$0.044  \\
			& & LiDAR+DVS & 0.091 $\pm$0.003 & 0.117 $\pm$0.004 & 0.054 $\pm$0.003 & 0.220 $\pm$0.023  \\
			
			& & LiDAR+RGB+DVS & 0.090 $\pm$0.004 & 0.110 $\pm$0.003 & 0.051 $\pm$0.003 & 0.215 $\pm$0.059  \\
			& \textbf{with Transformer} & LiDAR+RGB & 0.092 $\pm$0.003 & 0.110 $\pm$0.004 & 0.055 $\pm$0.003 & 0.226 $\pm$0.057  \\
			& & \textbf{LiDAR+DVS} & \textbf{0.085 $\pm$0.003} & \textbf{0.107 $\pm$0.002} & \textbf{0.051 $\pm$0.003} & \textbf{0.208 $\pm$0.032} \\
			
			\midrule
			\multirow{6}{*}{Evening}
			& & LiDAR+RGB+DVS & 0.088 $\pm$0.004 & 0.117 $\pm$0.003 & 0.047 $\pm$0.002 & 0.228 $\pm$0.031  \\
			& w/o Transformer & LiDAR+RGB & 0.090 $\pm$0.003 & 0.113 $\pm$0.004 & 0.048 $\pm$0.004 & 0.233 $\pm$0.055  \\
			& & LiDAR+DVS & 0.086 $\pm$0.003 & 0.102 $\pm$0.003 & 0.046 $\pm$0.002 & 0.219 $\pm$0.036  \\
			
			& & LiDAR+RGB+DVS & 0.087 $\pm$0.002 & 0.109 $\pm$0.003 & 0.046 $\pm$0.003 & 0.215 $\pm$0.041  \\
			& \textbf{with Transformer} & LiDAR+RGB & 0.089 $\pm$0.004 & 0.112 $\pm$0.003 & 0.047 $\pm$0.003 & 0.218 $\pm$0.047  \\
			& & \textbf{LiDAR+DVS} & \textbf{0.079 $\pm$0.003} & \textbf{0.100 $\pm$0.003} & \textbf{0.044 $\pm$0.002} & \textbf{0.211 $\pm$0.043} \\		
			
			\bottomrule
		\end{tabular}
	}
\end{table*}

\subsection{Ablation Analysis of Sensor Modalities and Fusion}
To rigorously isolate the contributions of our proposed upgrades, we conducted comprehensive ablation studies on both the sensory modalities and the fusion architecture, with the results detailed in Table \ref{tab:abl_results}. Specifically, the input configurations were systematically varied between traditional frame-based vision (LiDAR + RGB) and the asynchronous event stream (LiDAR + DVS) to evaluate their respective robustness against high-speed motion blur (sudden encounters) and degraded illumination. Concurrently, the fusion mechanism itself was ablated to compare the proposed dynamic cross-modal Transformer against standard static fusion operations. By analyzing these architectural permutations across both noon and evening conditions, we can empirically quantify the precise predictive gains attributable to each of our architectural design choices.

First, regarding sensor modalities, the combination of LiDAR + DVS significantly outperforms LiDAR + RGB. In both noon and evening conditions, the RGB camera suffers from exposure limitations and motion blur when a pedestrian suddenly enters the frame, degrading the latent features. The DVS, operating asynchronously, captures the microsecond-level pixel intensity changes of the moving pedestrian independent of lighting, allowing the model to react instantaneously. This is quantitatively reflected in the Response Time Delay (RTD) metric (see Table \ref{tab:abl_results}), where the LiDAR + DVS configuration achieves the lowest temporal lag, proving that asynchronous event streams inherently reduce perception latency. Interestingly, the LiDAR + DVS configuration also outperforms the tripartite LiDAR + RGB + DVS setup. This occurs because the inclusion of the RGB stream during high-speed sudden events introduces conflicting, blurry data into the fusion space. The network must expend representational capacity to filter out this visual noise, which slightly degrades the optimal policy compared to utilizing purely spatial (LiDAR) and purely dynamic (DVS) modalities.

Second, ablating the Transformer module highlights the necessity of dynamic cross-modal attention. DeepIPCv3 variants employing standard concatenation or recurrent feature fusion (without the Transformer attention blocks) exhibit higher steer and waypoint errors during sudden crossings. Static fusion mechanisms apply fixed weights to the input modalities, whereas the proposed Transformer module dynamically computes attention scores. When a pedestrian suddenly crosses the street, the cross-attention mechanism adaptively suppresses the static LiDAR features and maximizes the weight of the DVS event stream, enabling the highly reactive maneuverability demonstrated in our results.

\begin{table*}[htbp]
	\centering
	\caption{Comparative Evaluation Results}
	\label{tab:results}
	\resizebox{0.98\textwidth}{!}{%
		\begin{tabular}{ll cccccc}
			\toprule
			\multirow{2}{*}{\textbf{Condition}} & \multirow{2}{*}{\textbf{Model}} & \multicolumn{2}{c}{\textbf{Perception Task}} & \textbf{Planning Task} & \multicolumn{3}{c}{\textbf{Control Task}} \\
			\cmidrule(lr){3-4} \cmidrule(lr){5-5} \cmidrule(lr){6-8}
			& & \textbf{Seg. IoU $\uparrow$} & \textbf{Dep. MAE $\downarrow$} & \textbf{Waypoints ADE $\downarrow$} & \textbf{Steer MAE $\downarrow$} & \textbf{Cruise MAE $\downarrow$} & \textbf{RTD (s) $\downarrow$} \\
			
			\midrule
			\multirow{7}{*}{Noon}
			& Huang \textit{et al.} \cite{huang_model} & 0.742 $\pm$0.005 & - & - & 0.134 $\pm$0.002 & 0.077 $\pm$0.003 & 0.274 $\pm$0.061  \\
			& AIM-MT \cite{aim_mt} & 0.801 $\pm$0.004 & 0.082 $\pm$0.002 & 0.167 $\pm$0.004 & 0.120 $\pm$0.004 & 0.068 $\pm$0.003 & 0.259 $\pm$0.046  \\
			& LMDrive \cite{lmdrive} & - & - & \textbf{0.084 $\pm$0.002} & \textbf{0.104 $\pm$0.004} & 0.055 $\pm$0.003 & 0.244 $\pm$0.042  \\
			& TransFuser \cite{transfuser} & - & - & 0.133 $\pm$0.003 & 0.119 $\pm$0.002 & 0.066 $\pm$0.003 & 0.236 $\pm$0.040  \\
			& DeepIPC \cite{deepipc} & 0.805 $\pm$0.004 & - & 0.107 $\pm$0.003 & 0.127 $\pm$0.002 & 0.059 $\pm$0.004 & 0.240 $\pm$0.022  \\
			& DeepIPCv2 \cite{deepipcv2} & - & - & 0.093 $\pm$0.002 & 0.127 $\pm$0.004 & 0.053 $\pm$0.003 & 0.236 $\pm$0.028  \\
			& \textbf{DeepIPCv3} & \textbf{-} & \textbf{-} & 0.085 $\pm$0.003 & 0.107 $\pm$0.002 & \textbf{0.051 $\pm$0.003} & \textbf{0.208 $\pm$0.032}  \\
			
			\midrule
			\multirow{7}{*}{Evening}
			& Huang \textit{et al.} \cite{huang_model} & 0.799 $\pm$0.003 & - & - & 0.126 $\pm$0.003 & 0.073 $\pm$0.004 & 0.278 $\pm$0.049  \\
			& AIM-MT \cite{aim_mt} & 0.803 $\pm$0.004 & 0.070 $\pm$0.003 & 0.143 $\pm$0.002 & 0.115 $\pm$0.002 & 0.058 $\pm$0.004 & 0.254 $\pm$0.038  \\
			& LMDrive \cite{lmdrive} & - & - & \textbf{0.077 $\pm$0.004} & \textbf{0.099 $\pm$0.003} & 0.048 $\pm$0.005 & 0.240 $\pm$0.034  \\
			& TransFuser \cite{transfuser} & - & - & 0.115 $\pm$0.003 & 0.111 $\pm$0.002 & 0.058 $\pm$0.002 & 0.231 $\pm$0.057  \\
			& DeepIPC \cite{deepipc} & 0.804 $\pm$0.003 & - & 0.094 $\pm$0.002 & 0.122 $\pm$0.004 & 0.051 $\pm$0.003 & 0.236 $\pm$0.036  \\
			& DeepIPCv2 \cite{deepipcv2} & - & - & 0.088 $\pm$0.003 & 0.122 $\pm$0.002 & 0.047 $\pm$0.003 & 0.227 $\pm$0.045 \\
			& \textbf{DeepIPCv3} & \textbf{-} & \textbf{-} & 0.079 $\pm$0.003 & 0.100 $\pm$0.003 & \textbf{0.044 $\pm$0.002} & \textbf{0.211 $\pm$0.043} \\
			
			\bottomrule
		\end{tabular}
	}
	\vspace{-7pt}
\end{table*}

\subsection{Comparative Study against State-of-the-Art}

To benchmark DeepIPCv3, we compared its performance against several prominent end-to-end driving models, including Huang \textit{et al.} \cite{huang_model}, AIM-MT \cite{aim_mt}, TransFuser \cite{transfuser}, as well as our previous architectures, DeepIPC \cite{deepipc} and DeepIPCv2 \cite{deepipcv2}. As shown on Table \ref{tab:results}, the fully configured DeepIPCv3 achieves state-of-the-art performance, yielding the lowest ADE and MAE for waypoints, steer, and cruise controls. Models relying strictly on standard frame-based inputs, such as AIM-MT and Huang \textit{et al.}, exhibit the highest errors because they suffer from motion blur during sudden pedestrian crossings. This visual degradation forces these models to wait for subsequent frames to regain clear features, resulting in higher Response Time Delays (RTD) exceeding 0.25 seconds, which indicates a lag of at least one full frame (e.g., 0.274 s for Huang \textit{et al.}). Conversely, DeepIPCv3 utilizes the microsecond-level dynamic updates of the DVS to achieve the lowest overall RTD (0.208 s in noon conditions and 0.211 s in evening conditions). By keeping the RTD strictly under the 250 ms threshold, DeepIPCv3 consistently executes its evasive maneuver within the exact same frame window as the expert driver, proving its superior responsiveness. While TransFuser employs an attention mechanism to fuse multi-modal inputs, it remains bottlenecked by its reliance on the RGB stream. During sudden dynamic events, it attempts to fuse crisp LiDAR geometry with a blurry RGB frame, introducing visual noise into the latent representation. DeepIPCv3 effectively resolves this bottleneck by replacing the RGB stream with the asynchronous DVS event stream, providing microsecond-level dynamic updates and ensuring high-fidelity features.

To provide a more contemporary baseline, we additionally compared our model against LMDrive \cite{lmdrive}, a state-of-the-art Large Language Model (LLM)-based driving architecture. For a fair evaluation, we adapted LMDrive to our dataset by restricting the visual input to the front-view camera and generating the required navigation instructions from the local route points which is similar to what we did to Huang's model. Because our dataset lacks annotated hazard warnings, LMDrive's optional "Notice Instruction" module was omitted. As reflected in our results, LMDrive demonstrates exceptional spatial reasoning, achieving slightly lower waypoint prediction errors (ADE) than our proposed method, alongside highly comparable control predictions. However, DeepIPCv3 maintains a decisive advantage in the Response Time Delay (RTD) metric. This discrepancy highlights a fundamental difference in architectural philosophy: while LMDrive relies on explicit textual "Notice Instructions" (e.g., "Watch for pedestrians up front") to prime its attention for hazards, DeepIPCv3 achieves instantaneous reactivity intrinsically. By leveraging the DVS and the cross-modal Transformer, DeepIPCv3 inherently perceives and reacts to sudden dynamic encounters better than LMDrive when explicit warnings are absent. This proves that low-latency sensory fusion provides a safer foundation for emergency obstacle avoidance.

Notably, DeepIPC and DeepIPCv2, already outperform SOTA baselines in most evaluation metrics. This robust performance is fundamentally attributed to two key architectural choices seamlessly retained in DeepIPCv3. First, rather than forcing the network to perform raw end-to-end perception, we employ PolarNet to establish a robust, illumination-invariant structural prior of the drivable corridor. This explicitly decouples static scene understanding from dynamic obstacle avoidance. Second, our hybrid PID-MLP control stabilizes the vehicle's maneuverability, unlike purely regressive models that are prone to erratic waypoint predictions. The PID acts as a mathematically bounded safety net, ensuring smooth point-to-point trajectory tracking, while the MLP neural functions as a learned residual, providing the non-linear adjustments required for evasive maneuvering. DeepIPCv3 builds directly upon this stable foundation by upgrading the temporal feature fusion to a dynamic cross-modal Transformer. By learning to shift its attention between LiDAR and DVS, DeepIPCv3 enables the unprecedented reactivity during dynamic encounters.

\begin{figure*}
	\vspace{3pt}
	\begin{center}
		\includegraphics[width=\linewidth]{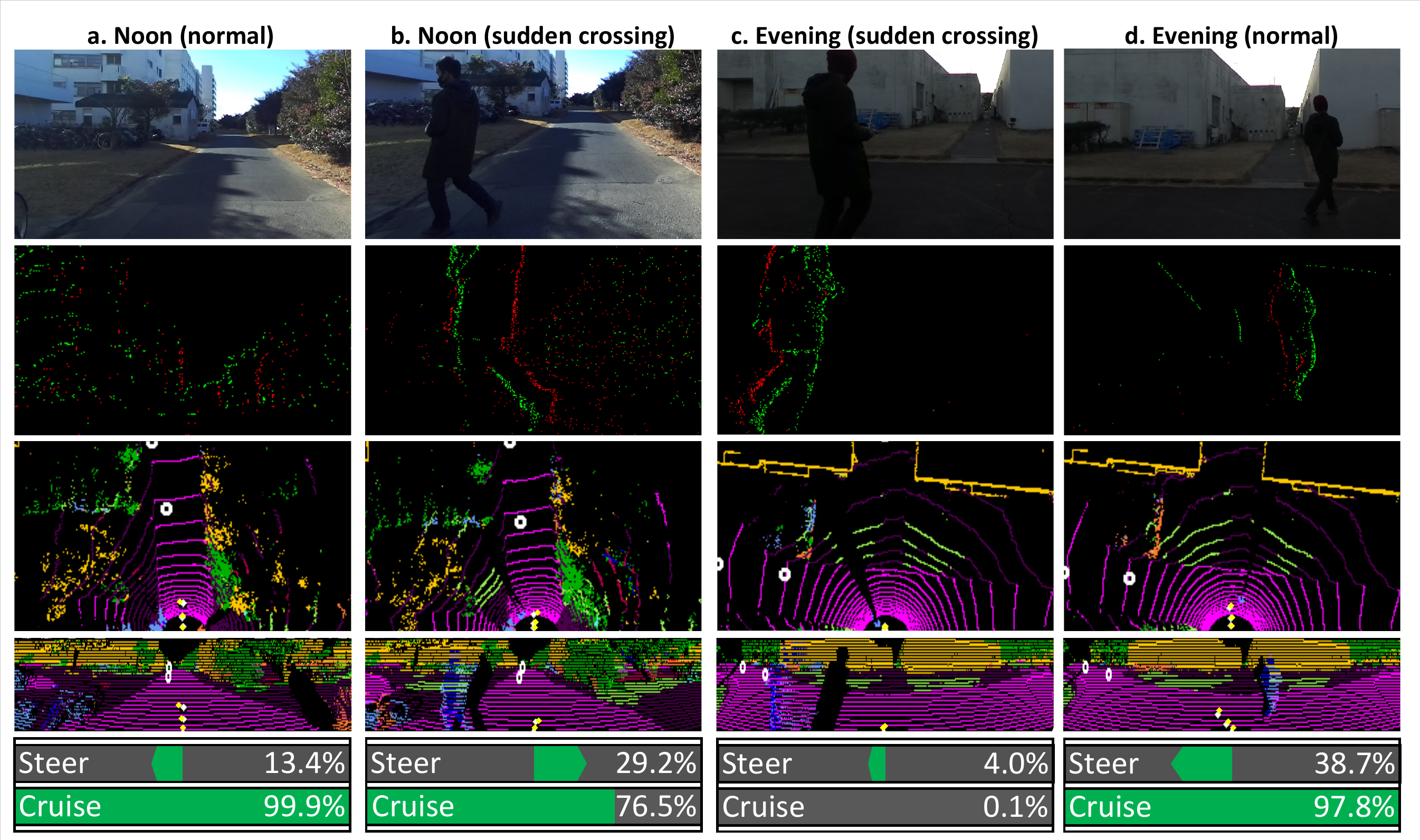}
		
	\end{center}
	\vspace{-6pt}
	\caption{Qualitative visualization of DeepIPCv3 navigating sudden pedestrian-crossing scenarios. Top to bottom: RGB (for monitoring only), DVS, LiDAR top view/BEV, LiDAR front view. Be noted that these LiDAR projected segmentation maps are concatenated with the its distance/depth maps projection. We omitted this detail for clarity. In the LiDAR images, expert ground-truth waypoints are shown in white, and the model's predicted waypoints are shown in yellow. Meanwhile, white hollow circles are route points projected to the local ego-vehicle coordinate. In noon conditions, DeepIPCv3 executes a reactive evasive steering maneuver to safely bypass the sudden pedestrian. From (a) to (b), during the trajectory correction following the route points (steer to the left), the model suddenly turns to the right, curving the predicted waypoints away from the pedestrian while applying a sharp steering override and partial braking. In illumination-degraded evening conditions, DeepIPCv3 executes a contextually safe, full-stop braking maneuver. From (c) to (d), the model decides to brake the vehicle to prevent a collision, the predicted waypoints compress directly in front of the vehicle, and the cruise command smoothly drops to zero. Then, it continues to the left following the route points after the pedestrian is not at the navigation path.}

	\label{fig:driving_rec} 
	\vspace{-7pt}
\end{figure*}

\subsection{Qualitative Results and Analysis}
To visually demonstrate the highly reactive capabilities of the proposed architecture, we present a qualitative analysis of the predicted local waypoints and control maneuvers during sudden pedestrian-crossing scenarios under varying illuminations as visualized in Fig. \ref{fig:driving_rec}.

\subsubsection{Performance in Noon Conditions}
During the noon evaluations, the environment is well-illuminated but highly dynamic. When a pedestrian rapidly enters the vehicle's path, DeepIPCv3 leverages the DVS event stream, which asynchronously captures the high-contrast moving edges of the pedestrian. The empirical trajectory outputs suggest the network instantaneously prioritizes the dynamic DVS updates. Quantitatively, the proposed method successfully handles the crossing pedestrian by achieving the lowest Response Time Delay (RTD of 0.208 s). Because the network reacts strictly within the initial 250 ms frame window, the model generates a highly active evasive maneuver without falling victim to the motion-blur delays that paralyze standard cameras. As observed in the qualitative frames, the predicted waypoints (yellow) curve away from the pedestrian, and the network successfully executes an immediate steering override combined with partial braking to safely navigate around the sudden dynamic obstacle that perfectly aligns with the expert demonstration (white).

\subsubsection{Performance in Evening Conditions}

The sudden crossing scenarios in evening conditions present a compounded challenge: simultaneous high-speed dynamic motion (requiring sub-second reaction times) and limited illumination. Despite these conditions, DeepIPCv3 remains highly stable. The LiDAR point cloud provides an illumination-invariant structural prior of the drivable corridor, while the exceptionally high dynamic range of the DVS allows the model to capture the moving human. In these specific evening scenarios, rather than attempting a high-speed evasion in low visibility, the network's learned policy prioritizes maximum safety by executing a full halt. Once again, the model demonstrates a highly responsive RTD of 0.211 s, proving that the DVS intensity updates allow the network to trigger the hard-braking sequence before frame-based models can reliably register the pedestrian's silhouette. Since the surrounding darkness obscures potential peripheral hazards, attempting a sharp evasive swerve would introduce an unacceptable risk of secondary collisions. The predicted waypoints compress directly in front of the vehicle, perfectly matching the expert's hard-braking demonstration to let the pedestrian cross safely. This explicitly demonstrates that the Transformer-based fusion of LiDAR and DVS not only decouples the system's responsiveness from ambient lighting but also enables contextually appropriate reactive behaviors.

\subsubsection{Limitations}
It is important to note that the efficacy of these reactive behaviors in both conditions is inherently sensitive to the choice of the micro-temporal window size, $\Delta t$, used for event aggregation. In this study, $\Delta t$ was strictly aligned with the 4 Hz sampling frequency of the LiDAR and the control loop ($\Delta t = 250$ ms) to maintain exact cross-modal synchronization. The optimal value of $\Delta t$ represents a critical trade-off: a window that is too small yields sparse, uninformative event tensors, whereas a window that is too large reintroduces motion blur into the event space and inherently delays actuation. Therefore, while a fixed 250 ms window proved optimal for our specific hardware frequencies and the observed pedestrian speeds, dynamically adapting $\Delta t$ based on the ego-vehicle's velocity and the scale of environmental dynamics remains a vital consideration for future optimization.
	
\section{Conclusions}

We proposed and evaluated DeepIPCv3, a novel multi-modal autonomous navigation framework designed to safely handle sudden pedestrian crossing scenarios, such as sudden pedestrian crossings. Through rigorous offline evaluations across well-illuminated noon and illumination-degraded evening conditions, the framework demonstrated state-of-the-art predictive accuracy. Our extensive comparative and ablation studies confirmed that fusing dense 3D LiDAR spatial geometry with asynchronous event streams from a Dynamic Vision Sensor (DVS) effectively eliminates the motion blur and exposure failures inherent to traditional RGB cameras. Furthermore, the introduced Transformer-based cross-attention mechanism proved crucial; rather than statically weighting inputs, it dynamically prioritizes high-speed DVS updates the moment a pedestrian appears. Qualitatively, DeepIPCv3 exhibited contextually optimal reactive behaviors, executing rapid evasive steering maneuvers to bypass obstacles during the noon evaluations, while prioritizing maximum safety by executing full-stop braking in the evening conditions. 

Despite these robust offline evaluations, future work must bridge the gap between offline policy validation and safe physical deployment. We plan to integrate the DeepIPCv3 architecture into high-fidelity closed-loop simulation platforms equipped with active event-camera and LiDAR plugins, allowing us to test dynamic stability and hybrid control execution without physical risk. Additionally, to accommodate the strict processing latency requirements of real-world autonomous driving, we aim to optimize the Transformer fusion module for edge-computing environments. By exploring network quantization and knowledge distillation, we intend to ensure that the model's ultra-low-latency perception translates seamlessly into real-time physical actuation on our robotic platform.

{\small
	\bibliographystyle{IEEEtran}
	\bibliography{references}
}

\begin{IEEEbiography}[{\includegraphics[width=1in,height=1.25in,clip,keepaspectratio]{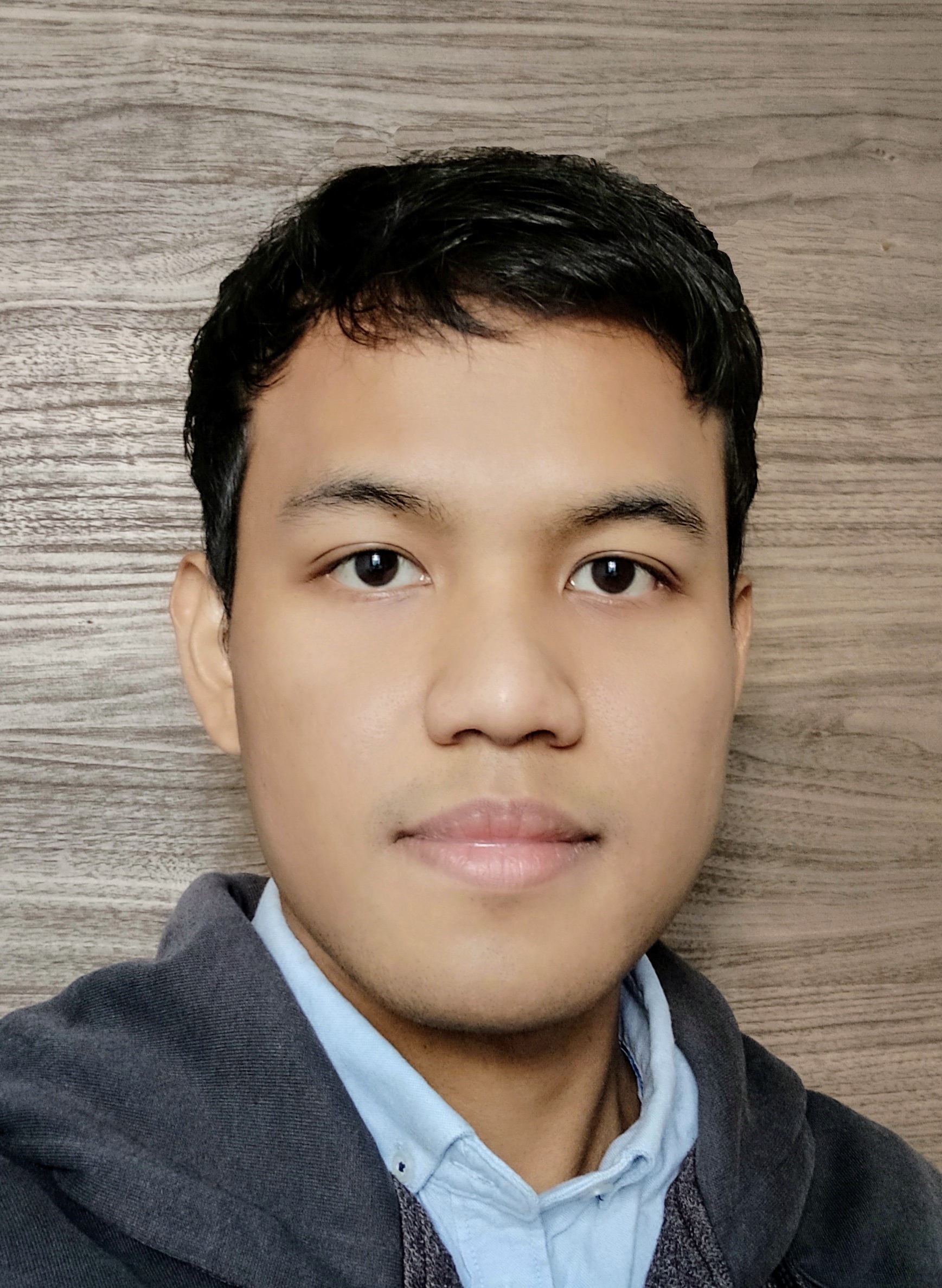}}]{Oskar Natan}
	(Member, IEEE) received his B.A.Sc. degree in Electronics Engineering and M.Eng. degree in Electrical Engineering from Politeknik Elektronika Negeri Surabaya, Indonesia, in 2017 and 2019, respectively. In 2023, he received his Ph.D. degree in Computer Science and Engineering from Toyohashi University of Technology, Japan. Since January 2020, he has been affiliated with the Department of Computer Science and Electronics, Universitas Gadjah Mada, Indonesia, first as a Lecturer and currently serves as an Assistant Professor. He has been serving as a reviewer/TPC member for some reputable journals and conferences. His research interests lie in the fields of deep learning, sensor fusion, robot vision, and hardware acceleration for various end-to-end systems. 
\end{IEEEbiography}

\vspace{-10mm}
\begin{IEEEbiography}[{\includegraphics[width=1in,height=1.25in,clip,keepaspectratio]{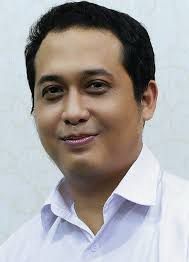}}]{Andi Dharmawan}
	(Member, IEEE) received his B.Sc. degree in Electronics and Instrumentation in 2006, M.Sc. degree in Computer Science in 2009, and Ph.D. degree in Computer Science in 2017, all from Universitas Gadjah Mada, Indonesia. From 2007 to 2009, he worked as a Research Assistant at the Department of Physics, Universitas Gadjah Mada. Since 2009, he has been a faculty member at the same department and has now become an Associate Professor at the Department of Computer Science and Electronics, Universitas Gadjah Mada. His research interests include intelligent control systems for robotics, autonomous unmanned systems, advanced control system development, and internet of things. 
\end{IEEEbiography}
\vspace{-10mm}

\begin{IEEEbiography}[{\includegraphics[width=1in,height=1.25in,clip,keepaspectratio]{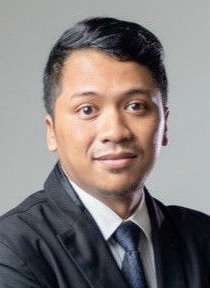}}]{Aufaclav Zatu Kusuma Frisky}
	(Member, IEEE) received his B.Sc. degree in Electronics and Instrumentation in 2012 from Universitas Gadjah Mada, Indonesia and his M.Sc. degree in Computer Science and Information Engineering in 2015 from National Central University, Taiwan. In 2022, he received his Dr.techn. degree in Informatics from Technische Universitat Wien (TU Wien), Austria. Since 2016, he has been affiliated with the Department of Computer Science and Electronics, Universitas Gadjah Mada, Indonesia, where he currently serves as an Assistant Professor. His research interests lie in the fields of computer vision, machine vision, robotic vision, image processing, pattern recognition, and artificial intelligence.
\end{IEEEbiography}
\vspace{-10mm}
\begin{IEEEbiography}[{\includegraphics[width=1in,height=1.25in,clip,keepaspectratio]{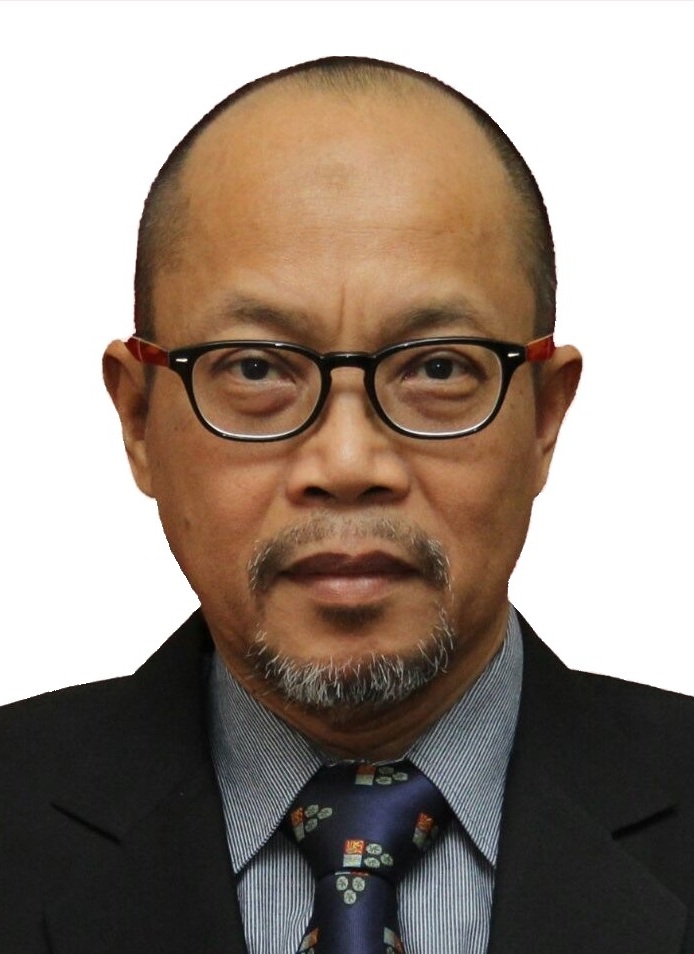}}]{Jazi Eko Istiyanto}
	(Member, IEEE) received his B.Sc. degree in Nuclear Physics in 1986 from Universitas Gadjah Mada, Yogyakarta, Indonesia. Then, he received his Postgraduate Diploma in Computer Programming and Microprocessors Applications in 1987, M.Sc. degree in Computer Science in 1988, and Ph.D. degree in Electronic Systems Engineering in 1995 from the University of Essex, United Kingdom. In 2010, Jazi became a full professor of Electronics and Instrumentation at the Department of Computer Science and Electronics, Universitas Gadjah Mada. He was also the Chairman of BAPETEN (Indonesia Nuclear Energy Regulatory Agency) from February 2014 until October 2021. He is also a registered engineer (electronic engineering) in Indonesia and the ASEAN countries. His research interests include embedded systems and cyber-physical systems security.
\end{IEEEbiography}


\vspace{-10mm}

\begin{IEEEbiography}[{\includegraphics[width=1in,height=1.25in,clip,keepaspectratio]{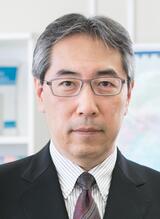}}]{Jun Miura}
	(Member, IEEE) received his B.Eng. degree in Mechanical Engineering and his M.Eng. and Dr.Eng. degrees in Information Engineering from the University of Tokyo, Japan, in 1984, 1986, and 1989, respectively. From 1989 to 2007, he was with the Department of Computer-controlled Mechanical Systems, Osaka University, Japan, first as a Research Associate and later as an Associate Professor. From March 1994 to February 1995, he served as a Visiting Scientist at the Department of Computer Science, Carnegie Mellon University, USA. In 2007, he became a Professor at the Department of Computer Science and Engineering, Toyohashi University of Technology, Japan, where he remains to the present. To date, he has received plenty of awards and authored or co-authored more than 265 peer-reviewed scientific articles in the field of robotics and autonomous systems in internationally reputable journals and conferences.
\end{IEEEbiography}

\vfill
\end{document}